\documentclass[10pt,twocolumn,letterpaper]{article}

\usepackage{iccv}
\usepackage{times}
\usepackage{epsfig}
\usepackage{graphicx}
\usepackage{amsmath}
\usepackage{amssymb}
\usepackage{multirow}
\usepackage{caption}
\usepackage{subcaption}
\usepackage{tablefootnote}

\usepackage[breaklinks=true,bookmarks=false]{hyperref}

\iccvfinalcopy % *** Uncomment this line for the final submission

\def\iccvPaperID{} % *** Enter the ICCV Paper ID here

\ificcvfinal\fi
\begin{document}

%%%%%%%%% TITLE
\title{Two-stream Flow-guided Convolutional Attention Networks for Action Recognition}

\author{An Tran \qquad Loong-Fah Cheong\\
Department of Electrical \& Computer Engineering, National University of Singapore\\
{\tt\small an.tran@u.nus.edu \quad eleclf@nus.edu.sg}
%\and
%Loong-Fah Cheong\\
%Department of Electrical \& Computer Engineering, National University of Singapore\\
%{\tt\small an.tran@u.nus.edu \quad eleclf@nus.edu.sg}
}

\maketitle
%\thispagestyle{empty}

%%%%%%%%% ABSTRACT
\begin{abstract}
	This paper proposes a two-stream flow-guided convolutional attention networks for action recognition in videos. The central idea is that optical flows, when properly compensated for the camera motion, can be used to guide attention to the human foreground. We thus develop cross-link layers from the temporal network (trained on flows) to the spatial network (trained on RGB frames). These cross-link layers guide the spatial-stream to pay more attention to the human foreground areas and be less affected by background clutter. We obtain promising performances with our approach on the UCF101, HMDB51 and Hollywood2 datasets.
\end{abstract}

%%%%%%%%% BODY TEXT
\section{Introduction}
Human action recognition in video is an important and challenging problem in computer vision. Like many other computer vision problems, an effective visual representation of actions in video data is vital to deal with these problems.

Over the last decade, there is a great evolution of features for action recognition from short video clips \cite{Wang2013a,Karpathy2014,Simonyan2014,Tran2014a}. The research works can be roughly divided into two mainstreams. The first type of representation is \textit{hand-crafted} local features in combination with the \textit{Bag-of-Features} (BoFs) paradigm \cite{Laptev2008,Klaser2008,Wang2013a}. Probably the most successful approach of local features representation is to extract improved dense trajectory features \cite{Wang2013a} and deploy Fisher vector representation \cite{Perronnin2010}. The second approach is to utilize \textit{deep learning algorithms} to learn features automatically from data (\eg, RGB frames or optical flows) \cite{Simonyan2014,Karpathy2014,Tran2014a,Varol2016,Wang2016b}.
Probably the most successful approach of local features representation is to extract improved dense trajectory features \cite{Wang2013a} and deploy Fisher vector representation \cite{Perronnin2010}. High performances of neural network architectures have been recently reported on video action recognition, specially those of two-stream convolutional networks \cite{Simonyan2014,Wang2016b}.

\begin{figure}[t!]
	\centering
	\includegraphics[width=\linewidth]{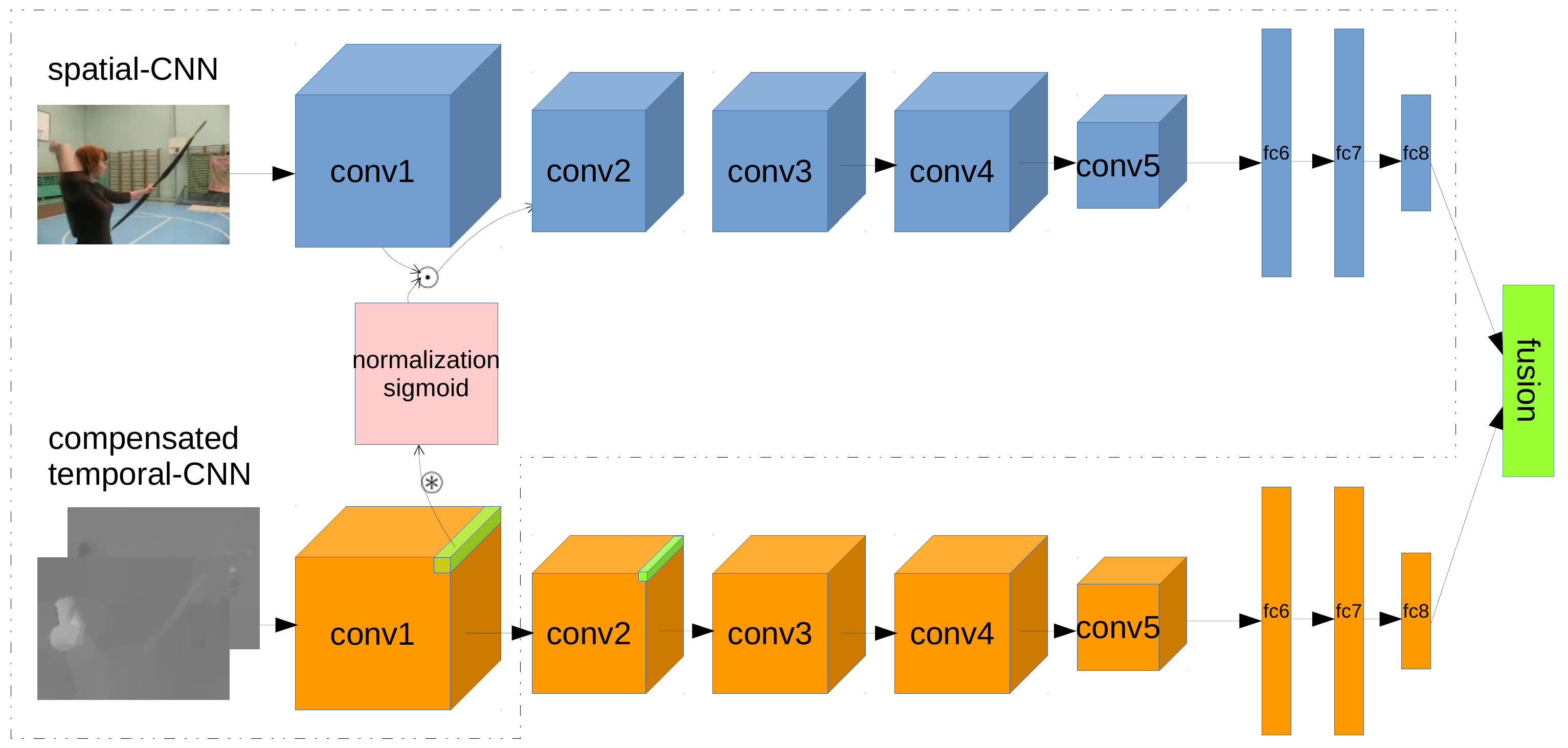}
	\caption{Our proposed \textit{Two-stream Flow-guided Convolutional Attention Networks} (Two-stream FCANs) for action recognition. A video clip is represented by RGB frames and optical flows. Two streams of data are fed into two separate CNNs: \textit{spatial stream} that models scene and object contexts (blue), while (compensated) \textit{temporal-stream} likely provides some motion-based attentions on foreground actions (orange). We leverage attentions provided from \textit{temporal-stream} to assist recognition processes in \textit{spatial-stream} by cross-link layers (pink). The attention weighted feature maps are fused by element-wise multiplication to preserve spatial-temporal structure in videos. The Two-stream FCAN refers to the entire architecture of two streams with the late fusion stage, whereas we call the area inside dashed lines the FCAN model. Best viewed in color.}
	\label{fig:main approach}
\end{figure}

\begin{figure}[t]
	\begin{center}
		\includegraphics[width=0.6\linewidth]{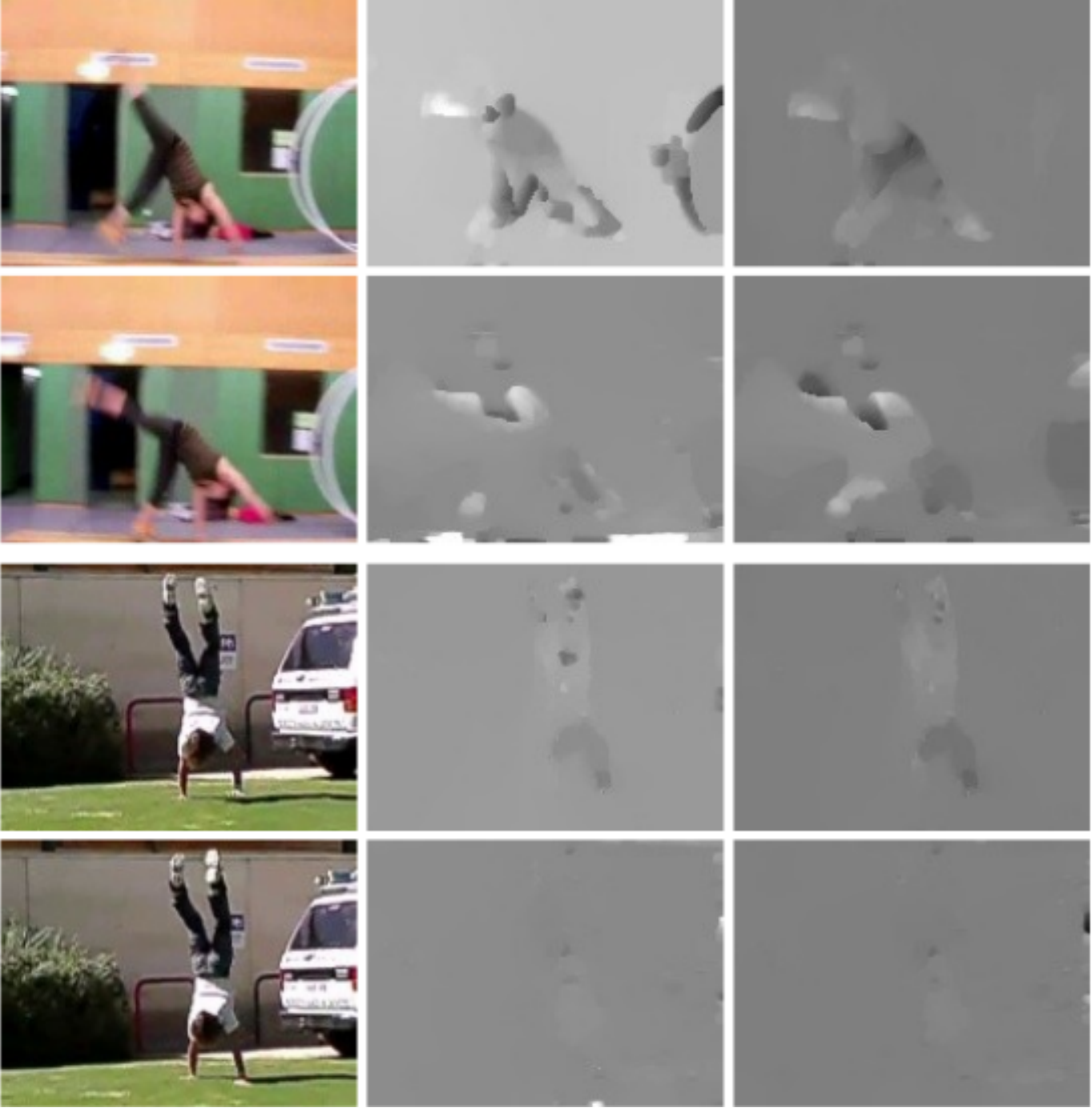}
	\end{center}
	\caption{Columns represent examples of three modalities of inputs: RGB frames, optical flows (\textit{x,y}-directions), and compensated optical flows. First two rows consist of two frames of a \textit{cartwheel} video in HMDB51 dataset, and last two ones consist of two frames of a \textit{handstand walking} video in UCF101 dataset.}
	\label{fig:data samples}
\end{figure}

Due to different network architectures and types of data in these two-stream networks, the learned features should have characteristics that help to deal with the different types of nuances in these specific data streams. RGB frames in video usually provide scene and object contexts in the background together with the human forms in the foreground. However, the spatial area occupied by the human foreground is usually much smaller than the area of the background such that it might not be effectively represented. On the contrary, optical flows in videos, when properly compensated for camera motion, immediately isolate the moving human silhouettes (see Figure~\ref{fig:data samples}), and provide motion cues. In view of the preceding discussion, feature responses of a CNN model (\ie, called \textit{spatial-CNN}) on RGB data are likely to be activations more on background contexts rather than on human foreground actions. In contrast, a CNN model (\ie, called \textit{temporal-CNN}) trained on flow data would often fire more on the human forms and movements. 

In this paper, we propose a novel flow-guided convolutional attention networks (FCANs) for action recognition based on the aforementioned two-stream network architecture (see Figure~\ref{fig:main approach}). This attention guiding is partly motivated by the primate visual system, whereby it is known that there is connectivity between the motion pathway and the form pathway \cite{VanEssen1983}. To model the attention guidance, we propose cross-link layers from the temporal stream to the spatial stream. There can be multiple such cross-link layers, but as we shall show later, the optimal number is in the range of one to two layers. Each cross-link layer has three components: (1) a convolution layer to reduce the dimension of the flow feature tensor; (2) a mean-variance normalization layer; (3) a sigmoid function. The final cross-link output is an attention map, which is used to control the level of activation in the corresponding layer in the spatial-stream via an element-wise multiplication.

Our contributions are two-fold. First, we propose a flow-guided convolution based attention mechanism for action recognition task. Second, we perform comprehensive evaluations two-stream FCANs based on 3d-convolution operations; we also explore the effects of different number of cross-link layers to understand where they are most effective. We visualize attentions provided by cross-link layers in our FCAN model (3D version) to show the attentive capacity of the (compensated) flows. Lastly, we achieve promising results on the HMDB51 and the UCF101 datasets. All codes and models \footnote{\url{https://github.com/antran89/two-stream-fcan}} are implemented in Caffe framework \cite{Jia2014}.

\section{Related Work}
\textbf{Visual features.} Many hand-crafted features have been proposed in the history of action recognition community, such as HOG/HOF \cite{Laptev2008}, HOG3D \cite{Klaser2008}, MBH \cite{Wang2013a}, \etc.

Inspired by recent successes in image classification \cite{Krizhevsky2012}, there have been extensions of the neural networks to the video action recognition problem \cite{Simonyan2014,Karpathy2014,Tran2014a}. CNN architectures play significant roles in these works, either as an individual module or as an encoder module for a type of recurrent neural networks (RNNs). Karpathy \etal \cite{Karpathy2014} propose a large-scale video dataset, namely Sport1M, and investigate different ways to embed temporal information into the current CNN architecture. Two-stream CNN model \cite{Simonyan2014} has demonstrated good performance on the UCF101 dataset \cite{Soomro2012} by combining predictions from two CNNs: \textit{spatial-CNNs} trained from RGB frames and \textit{temporal-CNNs} trained from optical flows. Recently, Tran \etal \cite{Tran2014a} extend 2D CNNs to 3D CNNs by developing 3d-convolution and 3d-pooling layers. In \cite{Ng_2015_CVPR,Donahue2015}, the output of a CNN's last layer is fed into a recurrent sequence model usually formed by LSTM cells. Interesting works \cite{Varol2016,Wang2016b} have attempted to model longer temporal information of videos.

\textbf{Attention for action recognition.} Attention is a mechanism used to confer more weights on a subset of features. The attention mechanism has also been applied to action recognition \cite{Li2016a,Sharma2015,Ballas2015,Bazzani2016}. Bazzani \etal use additional human fixation data to train mixture density network for saliency prediction and apply it to action recognition with the so called C3D \cite{Tran2014a} features obtained from the 3D CNNs. On the contrary, our approach does not use any additional data except flows to predict attentions. Furthermore, Sharma \etal \cite{Sharma2015} extract image features in each frame with the VGG16 CNN model \cite{Simonyan2015} and predict visual attention in each frame using a recurrent model with LSTM cell. The work most similar to ours is VideoLSTM \cite{Li2016a}. VideoLSTM \cite{Li2016a} uses convolutional LSTM trained on optical flow to predict attention for a second convolutional LSTM layer. Our FCAN is a convolution-based network that embeds attention in the process of action classification.

Before the deep learning era, there have been works incorporating saliency into action recognition from videos. Several saliency measures have been proposed for actions in \cite{Sultani2014,YeLuo_ICCV_2015} and they show improvements in the recognition accuracy when focusing attention on the foreground.

\section{Flow-guided Convolutional Attention Networks (FCANs)}
% just write it, and improve it latter, we need to have something for CVPR2017
% I will try to make a submission to the conference
We propose the cross-link layers to model the interactions of the two networks in the two-stream convolutional network \cite{Simonyan2014}. The whole network is differential, so it can be trained end-to-end with the stochastic gradient descent (SGD) and back-propagation algorithm \cite{lecun_1998}. The overall architecture of the FCAN is shown in Figure~\ref{fig:main approach}. In the ensuing discussion, we describe the 3D-FCAN model, which is the 3D version based on 3D convolution (\eg, C3D \cite{Tran2014a}) building blocks (please refer to the Supplementary Materials for the 2D-FCAN model based on 2D convolution (\eg, Alexnet \cite{Krizhevsky2012})).

\subsection{3D version of flow-guided convolutional attention networks}
With the assumption that the magnitudes of optical flows, when appropriately compensated for camera motions, usually correlate with the foreground regions, we develop the framework of flow-guided convolutional attention networks (FCANs) as shown in Figure~\ref{fig:main approach}. The C3D network \cite{Tran2014a} provides explicit representation of the time dimension in the architecture. Both the 3D-FCAN and 2D-FCAN have similar structures, except for the operations in the convolution and mean-variance normalization layer. Let $x\_rgb^l \in \mathbb{R}^{C_l \times T_l \times H_l \times W_l}$, $x\_flow^l \in \mathbb{R}^{C_l \times T_l \times H_l \times W_l}$ be the feature map of layer $l \in \{0,1,...,L\}$ in the \textit{spatial-} and \textit{temporal-C3D} respectively, with $C_l$, $T_l$, $H_l$, $W_l$ being the number of channels, temporal length, height and width of the feature map. Specifically, in the proposed FCANs, $x\_rgb$ and $x\_flow$ are the feature maps from a pooling layer in the C3D network. We develop attentive cross-link layers between the early pooling layers from the \textit{temporal-C3D} to the \textit{spatial-C3D}. As we shall show later, the optimal number of cross-link layers is between one and two, because the activations from the \textit{temporal-C3D} at these early stages are still largely retinotopic. They directly point to the foreground regions and help the \textit{spatial-C3D} learn distributed feature representation focused around these regions for the label prediction task. In the following, we report results for the case of only one attentive cross-link layer. Such cross-link layer includes the following three steps: reducing dimensions of a flow feature tensor $x\_flow^l$ (Equ.~\ref{equ:3d flow feature dim reduction}), mean-variance normalization (Equ.~\ref{equ:3d mean variance normalization on 3d-volume}), and attention prediction (Equ.~\ref{equ:2d attention scores}). We use a 3d-convolutional layer to reduce a flow feature tensor $x\_flow^l \in \mathbb{R}^{C_l \times T_l \times H_l \times W_l}$ to $x\_link^l \in \mathbb{R}^{1 \times T_l \times H_l \times W_l}$:
\begin{equation}
x\_link^l = W_{3D\_link} \circledast x\_flow^l.
\label{equ:3d flow feature dim reduction}
\end{equation}
where $\circledast$ is a 3d-convolution operation along the channel dimension $C_l$. We initialize the filter weights $W_{3D\_link}$ to $\dfrac{1}{C_l}$ in the training phase. Then, we normalize the feature tensor $x\_link^l$ by the mean $\mu$ and variance $\sigma$ of all the spatial-temporal feature activations in $x\_link^l$:
\begin{equation}
\hat{x}^l_{t,h,w} = \dfrac{x\_link^l_{t,h,w} - \mu}{\sigma}.
\label{equ:3d mean variance normalization on 3d-volume}
\end{equation}

The mean-variance normalization layer transforms the raw attention scores $x\_link^l$ into a normalized range $\hat{x}^l \in [-1, 1]$. Finally, the normalized attention score $\hat{x}^l$ is converted to an attention probability score $a^l \in [0, 1]$ by a sigmoid function:
\begin{equation}
a^l_{t,h,w} = \mathrm{sigmoid}(\hat{x}^l_{t,h,w}).
\label{equ:2d attention scores}
\end{equation}
where $a^l_{h,w} \in \mathbb{R}^{1 \times T_l \times H_l \times W_l}$.

We apply the flow-guided attention map on the feature map $x\_rgb^l$ of the \textit{spatial-C3D} by multiplicative interaction:
\begin{equation}
x\_rgb^l_{att} = \mathbf{r}(a^l, C_l) \odot x\_rgb^l.
\label{equ:2d spatial features attended}
\end{equation}
where $\mathbf{r}(a^l, C_l)$ is the $C_l$-times replication of the predictive attention map $a^l$ along the channel dimension, and $\odot$ denotes element-wise multiplication operation.

The attended feature map $x\_rgb^l_{att}$ is forwarded into the next layer $l+1$ to learn more abstract attended features:
\begin{equation}
x\_rgb^{l+1} = f_{spatial\_C3D}^{l+1}(x\_rgb^l_{att}).
\label{equ:2d spatial next features}
\end{equation}
where $f_{spatial\_C3D}^{l+1}$ is the operation in the next layer (\eg, convolution layer). Recall that we choose to have only one attentive cross-link layers because the activations in the higher layers of the \textit{temporal-C3D} would be more abstract and not necessarily correspond to the notions of foreground objects.

\section{Experiments}
\subsection{Data sets}
We evaluate the two-stream FCANs on three datasets for action recognition.

\textbf{UCF101 \cite{Soomro2012}.} This dataset is among the largest available action recognition benchmarks. UCF101 has 101 action classes and about 13320 videos (180 frames/video on average). There are three splits of training/testing data, and the performance is measured by mean classification accuracy across the splits.

\textbf{HMDB51 \cite{Kuehne2011}.} HMDB51 has 51 action categories and 6,766 videos. The dataset has two versions (original and motion-stabilized), and we use the original version which is more challenging for action recognition. The dataset has diverse background contexts and variations in motion pattern. It has three train/test splits with 3,570 training and 1,530 test videos.

\textbf{Hollwood2 \cite{Marszalek2009}.} The Hollywood2 \cite{Marszalek2009} dataset has 12 categories with 1,707 videos, which consist of 823 training and 884 test videos. The performance is measured by mean average precision (mAP) over all classes.

\subsection{Implementation details}
\label{sect: fcan implementation details}
\textbf{Video preprocessing.} For direct comparison with the two-stream CNNs work \cite{Simonyan2014}, we sample a fixed number of frames (\ie, 25) per video with equal temporal spacing in both training and testing. Optical flows are computed with TV-L1 \cite{Zach07aduality}. We choose an OpenCV implementation of TV-L1 because it has a good balance of efficiency and accuracy.

\textbf{Compensated flows.} Similar to the Improved Dense Trajectories work \cite{Wang2013a}, we deploy a global motion estimation method based on the assumption that two consecutive frames are related by a homography. Removal of this background motion (induced by the camera motion) renders the optical flow magnitudes more indicative of the locations of the human silhouettes (\eg, Figure~\ref{fig:data samples}). After compensation, we extract the $x-$, $y-$ optical flows and convert them into gray-scale images $[0, 255]$ by a linear rescaling. This rescaling has two-fold benefits. First, it will reduce the size of the flow datasets dramatically, as we now save the flow fields as images rather than as floating point numbers (\eg, from few TBs to dozen of GBs in UCF101). Second, by saving the flow fields as images, we are able to fine-tune our CNNs from models pre-trained with large-scale image dataset (\ie, ImageNet).

\textbf{ConvNet architectures.} We utilize the C3D \cite{Tran2014a} as the main component for our two-stream FCAN network. During our experiments, in the C3D network, we achieve higher performance by setting a high dropout ratio of 0.9 and 0.8 for fully connected layers fc6, fc8 respectively. We need to have more regularization (\ie, higher dropout) in the C3D network due to the higher risk of over-fitting for the high capacity C3D models when dealing with small datasets (\eg, UCF101, HMDB51). We also experiment with 2D-CNN architecture (\eg, AlexNet), but flow-guided attentions does not provide benefits to frame-based representations.

\textbf{Data augmentation.} At training time, we sample 25 (overlapping) clips per videos with a temporal length of 1 frame for the 2D-CNN and 16 frames for the C3D network. We also adopt the corner and multi-scale cropping strategy for training the baseline models \cite{Wang2016b}. However, for training the FCAN models, we do not use multi-scale cropping because the attention maps already delineate which regions require more resolution and which require more of an overall gist for background context. Note that in our case, each random crop sample should apply to the same location of both the RGB and flow images; without this correspondence, the cross-link layers would be meaningless.

\begin{figure}[t!]
	\centering
	\includegraphics[width=\linewidth]{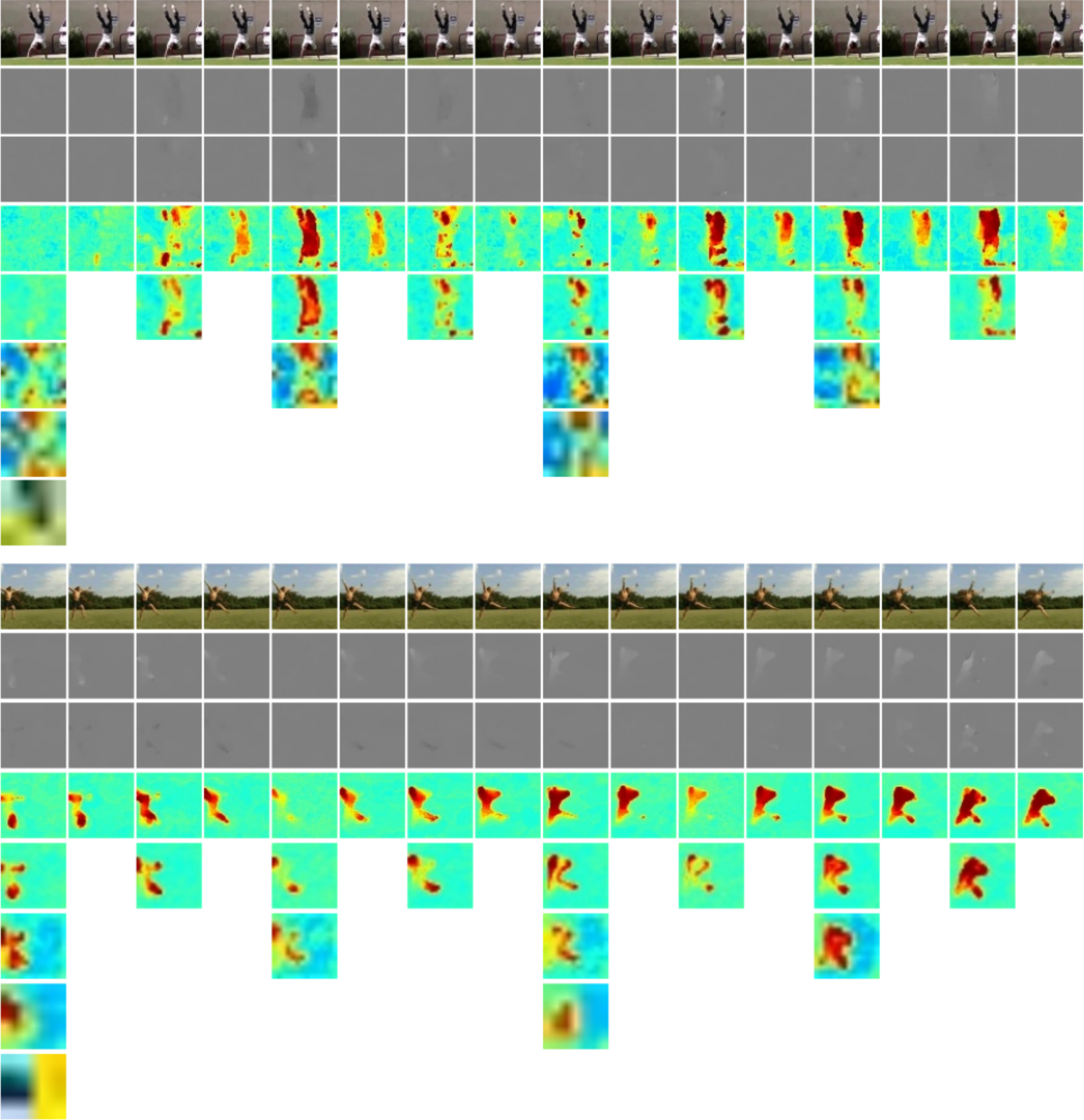}
	\caption{Visualizations of flow-guided attention provided by a temporal-C3D network. The top half shows a handstand walking video in UCF101, while the bottom one shows a cartwheel video in HMDB51 dataset. From top to bottom in each half: 16 RGB frames, flow-x, flow-y, attentions at layers pool1, pool2, pool3, pool4 and pool5. The spatial-temporal resolution of feature maps sequentially decreases with pooling layers, but we upsample the feature maps to have same sizes. Warm color indicates high saliency value. Best viewed in color.}
	\label{fig:temporal-c3d attention visualization}
\end{figure}

\textbf{Pre-trained weights.} In order not to overfit the CNN models in our experiments, we follow the initialization strategies in \cite{Wang2016b}. For the \textit{spatial-CNN} and \textit{spatial-C3D} networks, we initialize them with the pre-trained weights obtained from large-scale datasets (\ie, ImageNet \cite{Jia2014} and Sports1M \cite{Tran2014a} respectively).

\begin{table*}
	\begin{center}
		\begin{tabular}{|l|*{2}{c|}*{2}{c|}*{1}{c|}}
			\hline
			\multirow{2}{*}{\bfseries Models} & \multicolumn{2}{c|}{\bfseries UCF101} & \multicolumn{2}{c|}{\bfseries HMDB51} & \bfseries Hollywood2 \\
			\cline{2-6}
			& Clip acc. & Video acc. & Clip acc. & Video acc. & mAP		 \\
			&  (\%) &  (\%) &  (\%) &  (\%) &  (\%)  \\
			\hline\hline
			spatial-C3D  		 & 80.5 & 83.6  & 51.3 	& 53.9	&  43.6 \\
			temporal-C3D  		 & 70.6 & 83.1  & 38.5 	& 50.7	&  53.9 \\
			temporal-C3D-comp  	 & 72.0 & 84.6  & 42.6 	& 55.8	&  67.7 \\
			VideoLSTM RGB\cite{Li2016a}  & - & 79.6 & - & 43.3  & - \\
			VideoLSTM flow\cite{Li2016a} & - & 82.1 & - & 52.6  & - \\
			\hline
			FCAN		 & 81.5		& 85.4  & 51.6	& 54.6 		&  46.9\\
			FCAN-comp	 & 82.7		& 87.2  & 53.5	& 56.9 		&  50.3\\
			%			Early fusion		 & 81.7 & 85.3  & 52.3  & 55.3	\\
			\hline
			VideoLSTM two-stream \cite{Li2016a} 	& - 	& 89.2  & - &  56.4  & - \\
			Twostream-C3D 		 & 86.8 	& 91.8  & 54.8  & 64.4  &  51.2 \\
			Twostream-C3D-comp   & 86.8 	& 91.4  & 55.7  & 67.1  &  65.9 \\
			\hline
			Twostream-FCAN		 & \textbf{87.2}		& 91.9  & 54.8	& 63.3  &  56.3 \\
			Twostream-FCAN-comp	 & 86.7		& \textbf{91.9}  & \textbf{55.9}  & \textbf{68.2}  & \textbf{71.1} \\
			\hline
		\end{tabular}
	\end{center}
	\caption{Results for two-stream FCAN models and their baselines on UCF101 and HMDB51 (both split 1) dataset. 3D convolutional neural networks have inputs with temporal length of 16 frames for both the RGB and optical flow modalities. The two-stream FCAN network has one attentive cross-link layer.}
	\label{tab: 3D-FCAN on UCF101, HMDB51}
\end{table*}

\textbf{Training.} Our attention network is trained end-to-end with the standard back-propagation algorithms. We use the mini-batch stochastic gradient descent (SGD) algorithm to optimize the cross-entropy error function. The initial learning rate is 0.0001. We use mini-batches of 256 samples for the 2D-CNN networks, and 128 samples for the C3D architectures. For UCF101, we optimize the networks for 20K iterations, during which the learning rate is twice decreased with a factor of $0.1$ at the 12K and 18K iterations. Due to the smaller dataset size of HMDB51 and Hollywood2, we run the SGD algorithm for 10K iterations and reduce the learning rate with a factor of $0.1$ at the 4K and 8K iterations. In contrast to \cite{Tran2014a}, we do not train a SVM on features fc6 extracted from the C3D models and our models are trained and tested in an end-to-end fashion. As can be shown in Section~\ref{sect: Results and analysis}, the performance of our end-to-end training is better than the results of fc6+SVM pipeline reported in \cite{Tran2014a}.

\textbf{Testing.} For a fair comparison, we also adopt the same testing scheme in other CNN-based works (\eg, two-stream CNNs \cite{Simonyan2014}, temporal segment networks \cite{Wang2016b}). Given a test video, we sample 25 segments of RGB or flow frames with equal temporal spacing between them. For each segment, we crop the center of a frame to evaluate a model. The final score of the video is computed by averaging the scores across different crops and segments. We find that averaging the last fully connected layer (\ie, fc8) scores always produce better results than the softmax scores.

\subsection{Baselines}
We compare our two-stream FCAN models with a set of baselines proposed recently \cite{Simonyan2014,Li2016a}. The foremost baselines for our two-stream FCAN models are two-stream C3D. Besides, we also compare our two-stream FCAN with the following model:

\textbf{VideoLSTM \cite{Li2016a}.} VideoLSTM \cite{Li2016a} utilizes a convolutional LSTM to estimate motion-based attention. In contrast, we use convolution layers in a \textit{temporal-CNN} network to provide flow-based attention. The results of VideoLSTM are directly extracted from \cite{Li2016a}.

\subsection{Results and analysis}
\label{sect: Results and analysis}
This section reports the performances of our two-stream FCAN models, effects of compensated flows on the FCAN models, and the results of some exploratory studies.

\begin{figure*}[t!]
	\centering
	\begin{subfigure}[t]{0.5\textwidth}
		\centering
		\includegraphics[height=1.5in,keepaspectratio]{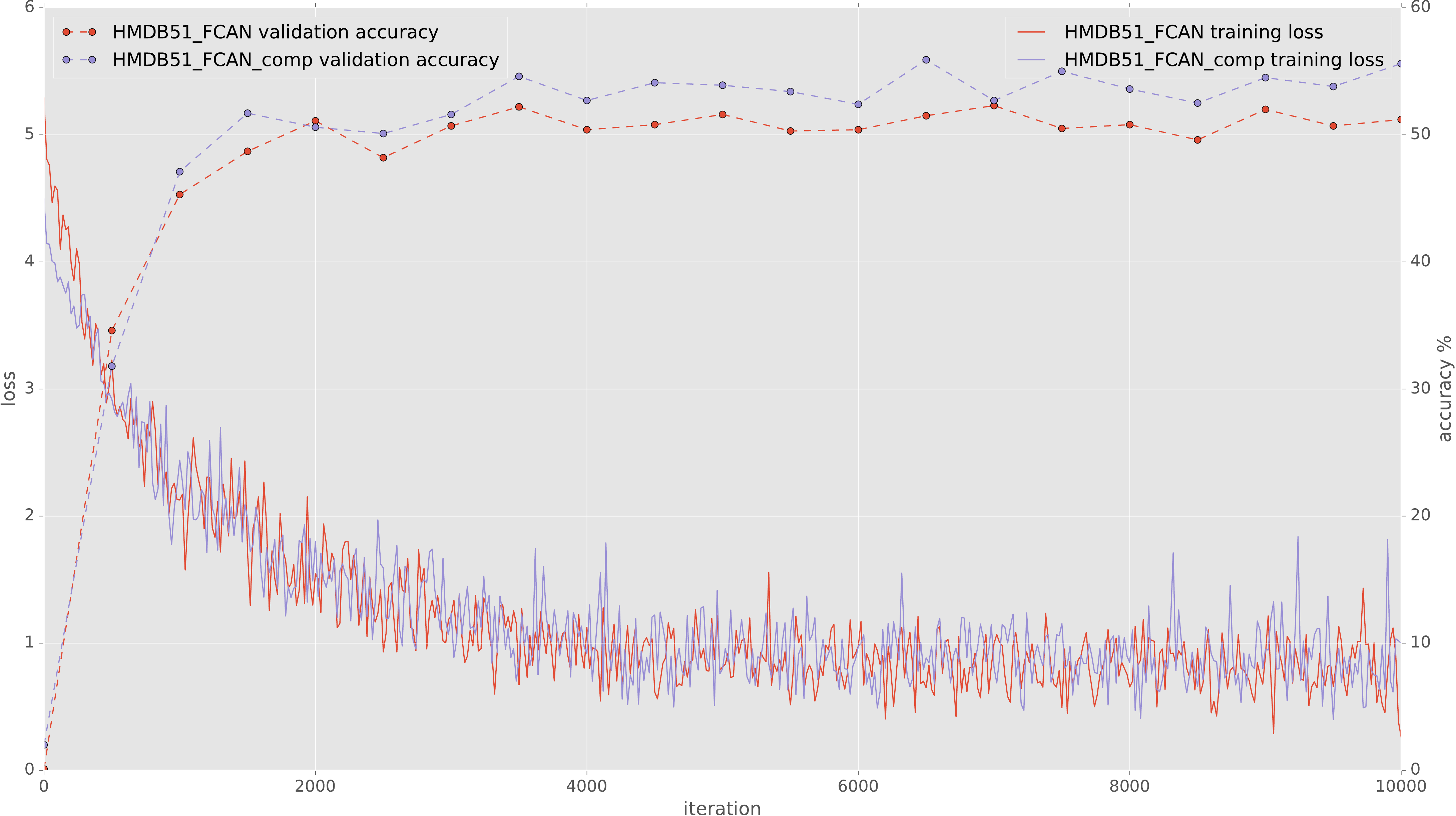}
		\caption{HMDB51}
	\end{subfigure}%
	~ 
	\begin{subfigure}[t]{0.5\textwidth}
		\centering
		\includegraphics[height=1.5in,keepaspectratio]{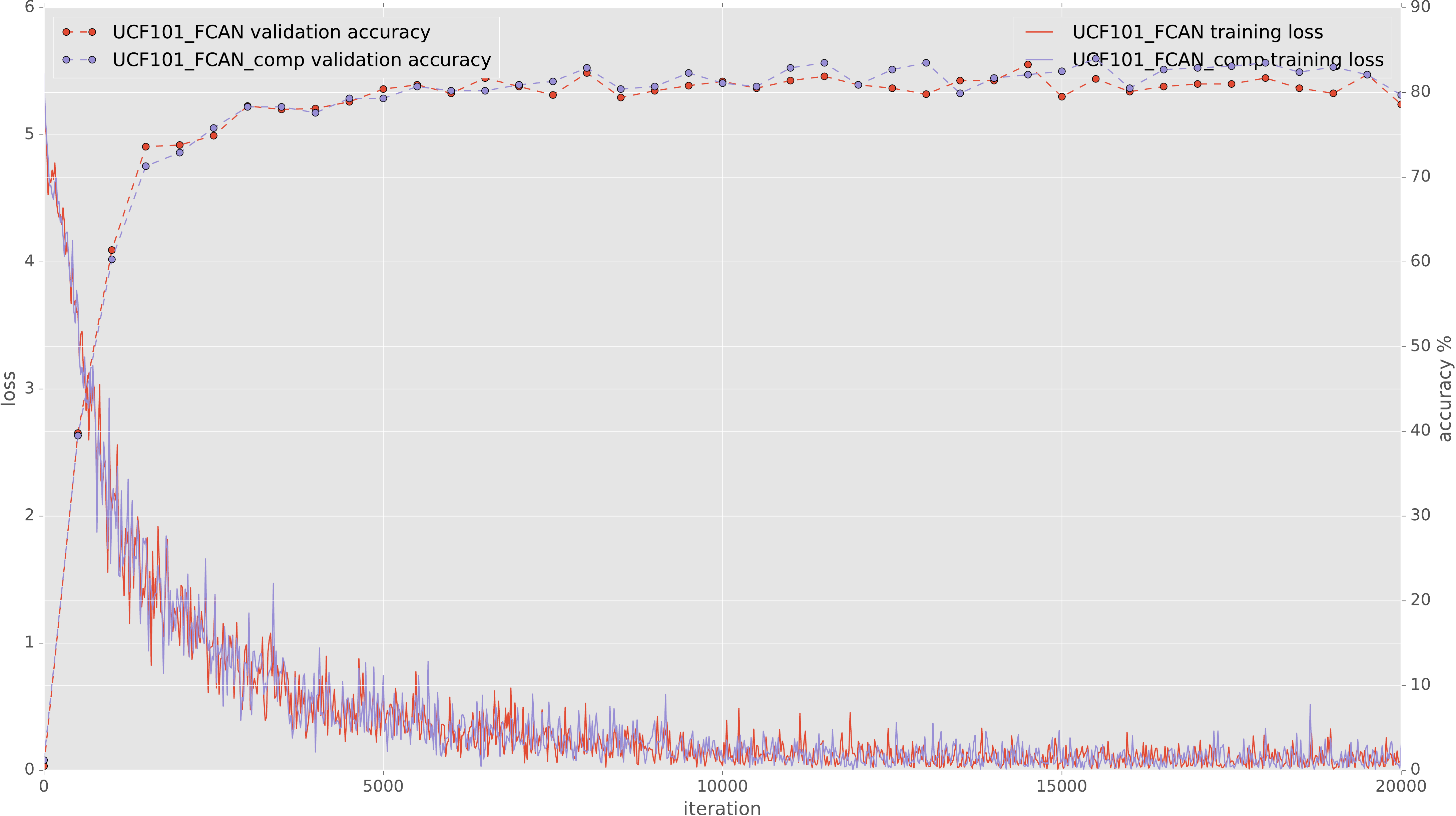}
		\caption{UCF101}
	\end{subfigure}
	\caption{Training loss and validation accuracy of FCAN and FCAN-comp models on UCF101 and HMDB51 (both split 1) dataset.}
	\label{fig: learning effects of FCAN-comp}
\end{figure*}

\textbf{Performance of two-stream FCAN networks.} Table~\ref{tab: 3D-FCAN on UCF101, HMDB51} shows performance of the two-stream FCAN on three datasets. With compensated flows, our two-stream FCAN demonstrates better performances than two baselines: two-stream C3D and videoLSTM two-stream \cite{Li2016a}. In particular, two-stream FCAN-comp (with compensated flows) outperforms two-stream C3D-comp 0.5\% on UCF101, 1.1\% on HMDB51 and 1.5\% on Hollywood2, and the performance gain over the baseline videoLSTM two-stream \cite{Li2016a} is much more significant: 2.7\% on UCF101 and 11.8\% on HMDB51. Focusing just on the FCAN networks (recall from Figure~\ref{fig:main approach} that FCAN is largely the spatial part of our architecture), we too observe consistent improvements over \textit{spatial-C3D} and ``videoLSTM RGB'' in terms of video-level accuracy.
Lastly, it is also evident that motion compensation is important in improving the performance of our FCAN networks. The improvement is more significant in HMDB51 than in UCF101 because many videos in HMDB51 contain more complex camera motions. Only with compensated flows, human foregrounds stand out from the background (\eg, Figure~\ref{fig:data samples}). Therefore, attentive effects in FCANs become more substantial. Figure~\ref{fig: learning effects of FCAN-comp} also shows that FCAN models learned on compensated flows have better generalization ability than on normal flows, especially on HMDB51 dataset.

\begin{table*}
	\begin{center}
		\begin{tabular}{|l|*{2}{c|}*{2}{c|}}
			\hline
			\multirow{2}{*}{\bfseries Models} & \multicolumn{2}{c|}{\bfseries UCF101} & \multicolumn{2}{c|}{\bfseries HMDB51} \\
			\cline{2-5}
			& \bfseries Clip acc. (\%) & \bfseries Video acc. (\%) & \bfseries Clip acc. (\%) & \bfseries Video acc. (\%) \\
			\hline\hline
			FCAN-comp pool1  		 & 82.7  & 87.2 &  53.5	 & 56.9		\\
			FCAN-comp pool2 		 & 82.1  & 86.3 &  52.7  & 57.7		\\
			FCAN-comp pool3 		 & 81.6	 & 86.1	&  52.0  & 57.1		\\
			FCAN-comp pool4 		 & 79.9	 & 86.2 &  48.2	 & 52.2		\\
			FCAN-comp pool5 		 & 78.3  & 85.0	&  45.1  & 49.1		\\
			\hline
		\end{tabular}
	\end{center}
	\caption{Evaluations of FCAN-comp with different numbers of cross-link layers on UCF101 (split 1) and HMDB51 (split 1) dataset. The suffix pool-n means that there are cross-link layers from layer 1 to layer n. 3D convolutional neural networks have inputs with temporal length of 16 frames for both the RGB and optical flow modalities. All flows in this experiment are compensated flows. }
	\label{tab: 3D-FCAN in different layers}
\end{table*}

\textbf{Exploration study.} We evaluate the effects of varying the number of cross-link layers from the lower to the higher layers. In particular, we start with just one cross-link layer at layer pool1, and then successively add more cross-link layers until all five layers are connected. Table~\ref{tab: 3D-FCAN in different layers} shows the performance of FCAN when we gradually increase the cross-link layers from the lower to higher layers. In general, the performance of FCAN-comp gradually decreases when the number of cross-link layers increases from one to five layers. FCAN-comp achieves peak performance at the first pooling layer pool1 in UCF101, while its peak performance in the HMDB51 is attained at adding the pooling layer pool2 (\ie, with 87.2\% and 57.7\% respectively in video accuracy). From the visualizations of the activation maps in Figure~\ref{fig:temporal-c3d attention visualization}, it can be seen that those in the higher layers (\eg, pool3, pool4, pool5) are no longer retinotopic, and may not correspond to the human silhouettes. Therefore, creating cross-link at these layers is counter-productive, usually degrading the performance of our classifiers.

\textbf{Visualization of attention layers.} In Figure~\ref{fig:temporal-c3d attention visualization}, we provide a visualization of the attention maps provided by the flows in Equ.~\ref{equ:2d attention scores}, using  two video sequences from the UCF101 and HMDB51 dataset. In the \textit{cartwheel} sequence, the motions of the actor are significant, and our attention maps in the pool1 and pool2 cross-link layers are indeed indicative of the actor's silhouettes. Attention map from the pool3 layer begins to be blurry. At higher layers (\eg, pool4, pool5), the attention maps have more abstract and complex patterns. Similar trends also appear in the \textit{handstand walking} sequence. These observations and the quantitative results in Table~\ref{tab: 3D-FCAN in different layers} corroborate our design choice of having one attentive cross-link layers.

\begin{table*}
	\begin{center}
		\begin{tabular}{|l|*{2}{c|}*{2}{c|}}
			\hline
			\multirow{2}{*}{\bfseries Models} & \multicolumn{2}{c|}{\bfseries UCF101} & \multicolumn{2}{c|}{\bfseries HMDB51} \\
			\cline{2-5}
			& \bfseries Clip acc. (\%) & \bfseries Video acc. (\%) & \bfseries Clip acc. (\%) & \bfseries Video acc. (\%) \\
			\hline\hline
			Twostream-TSN \cite{Wang2016b} \footnotemark & 81.3   & 91.5	 &  49.1 	& 64.5  \\
			Twostream-FCAN-comp		   		& 86.7   & 91.9  &  55.9  	& 68.2  \\
			\hline
			Twostream-FCAN-comp	+ Twostream-TSN	 & 88.9  & 93.4  &  61.3  	& 70.1  \\
			\hline
		\end{tabular}
	\end{center}
	\caption{Ensemble of TSN-BatchNorm-Inception \cite{Wang2016b} and FCAN features on UCF101 (split 1) and HMDB51 (split 1) dataset. 3D convolutional neural networks have inputs with temporal length of 16 frames for both RGB and optical flow modalities. All features are combined with equal weights.}
	\label{tab: Combination of different features}
\end{table*}

\begin{figure}[t!]
	\centering
	\includegraphics[width=\linewidth]{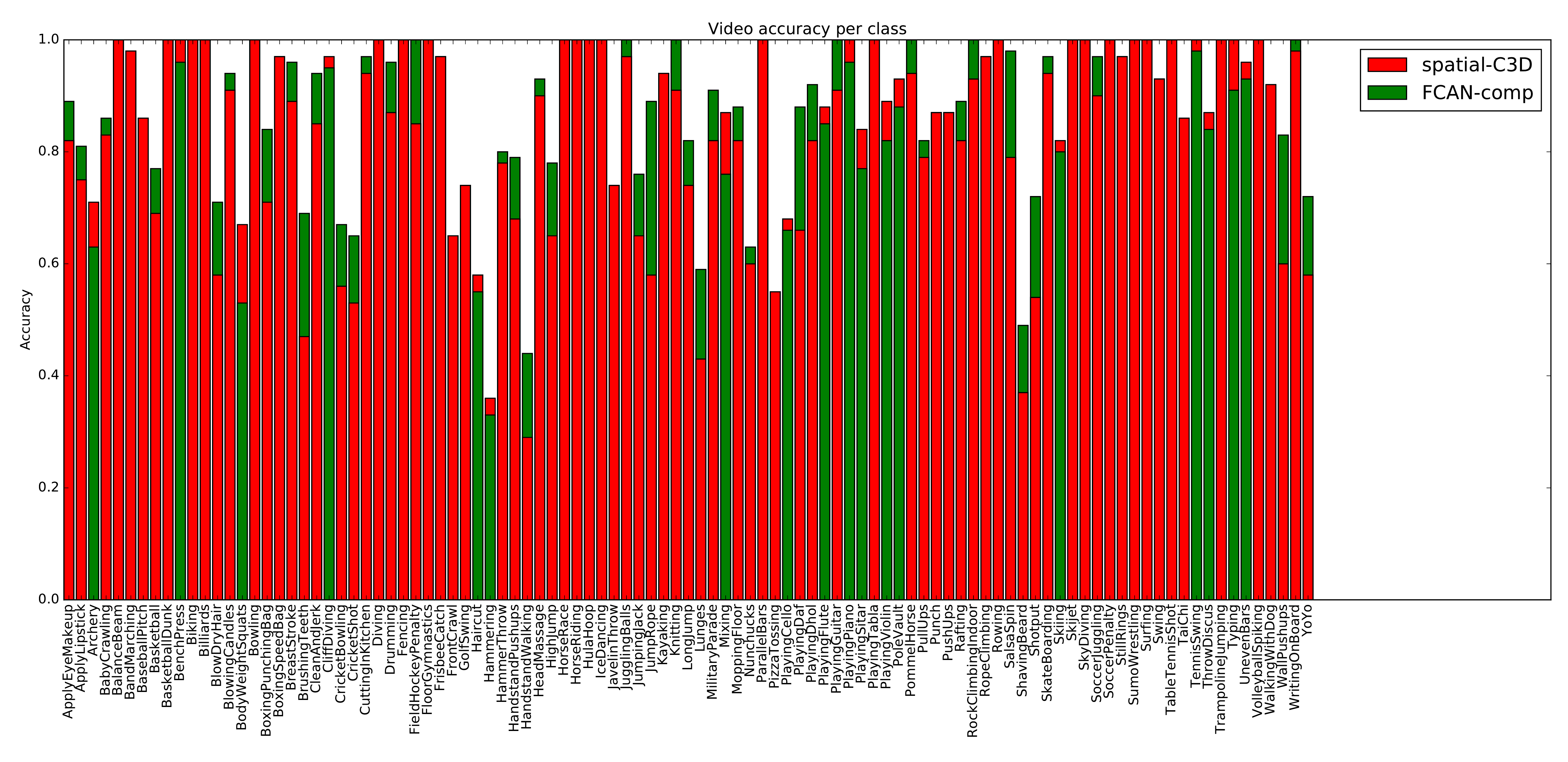}
	\caption{Classwise accuracy of FCAN-comp compared to spatial-C3D model. If a bar has only one color (red), both models have the same performance on the corresponding class. If green is on top of a bar, our FCAN-comp improves the accuracy, and vice versa.}
	\label{fig: ucf101 class accuracy}
\end{figure}

\textbf{Errors analysis.} Now, delving into the performance gain of FCAN over \textit{spatial-C3D}, we find that FCAN has better accuracy in all five action types in UCF101. Specifically, the performance gain is more noticeable on the action types of ``human-human interaction'', ``human-object interaction'' and ``body-motion only''. These action classes mostly tend to be those categories which have significant motions, allowing the compensated optical flows to pick up vividly the human form. In UCF101, some classes gain remarkable performance over \textit{spatial-C3D}, such as JumpingJack (76\% vs. 65\%), JumpRope (95\% vs. 58\%), HandstandWalking (44\% vs. 29\%), HandstandPushups (82\% vs. 68\%), Lunges (57\% vs. 43\%), MilitaryParade (94\% vs. 82\%), WallPushups (77\% vs. 60\%) and SalsaSpin (98\% vs 78\%) (see details in Figure~\ref{fig: ucf101 class accuracy}). FCAN obtains marginal improvements over \textit{spatial-C3D} on ``playing musical instruments'' because there are not much motions in the video sequences. In some ``sports'' sequences, FCAN's performance gain over spatial-C3D is also significant, such as Clean\&Jerk (97\% vs. 85\%), CricketBowling (72\% vs. 56\%), and CricketShot (63\% vs. 53\%). Scene contexts play important roles in sports sequences, and if the motions are also difficult to be picked up (\eg the swing of a golf club), then the improvement of FCAN over \textit{spatial-C3D} is limited compared to other types of actions.

\begin{figure}[t]
	\begin{center}
		\includegraphics[width=0.90\linewidth,keepaspectratio]{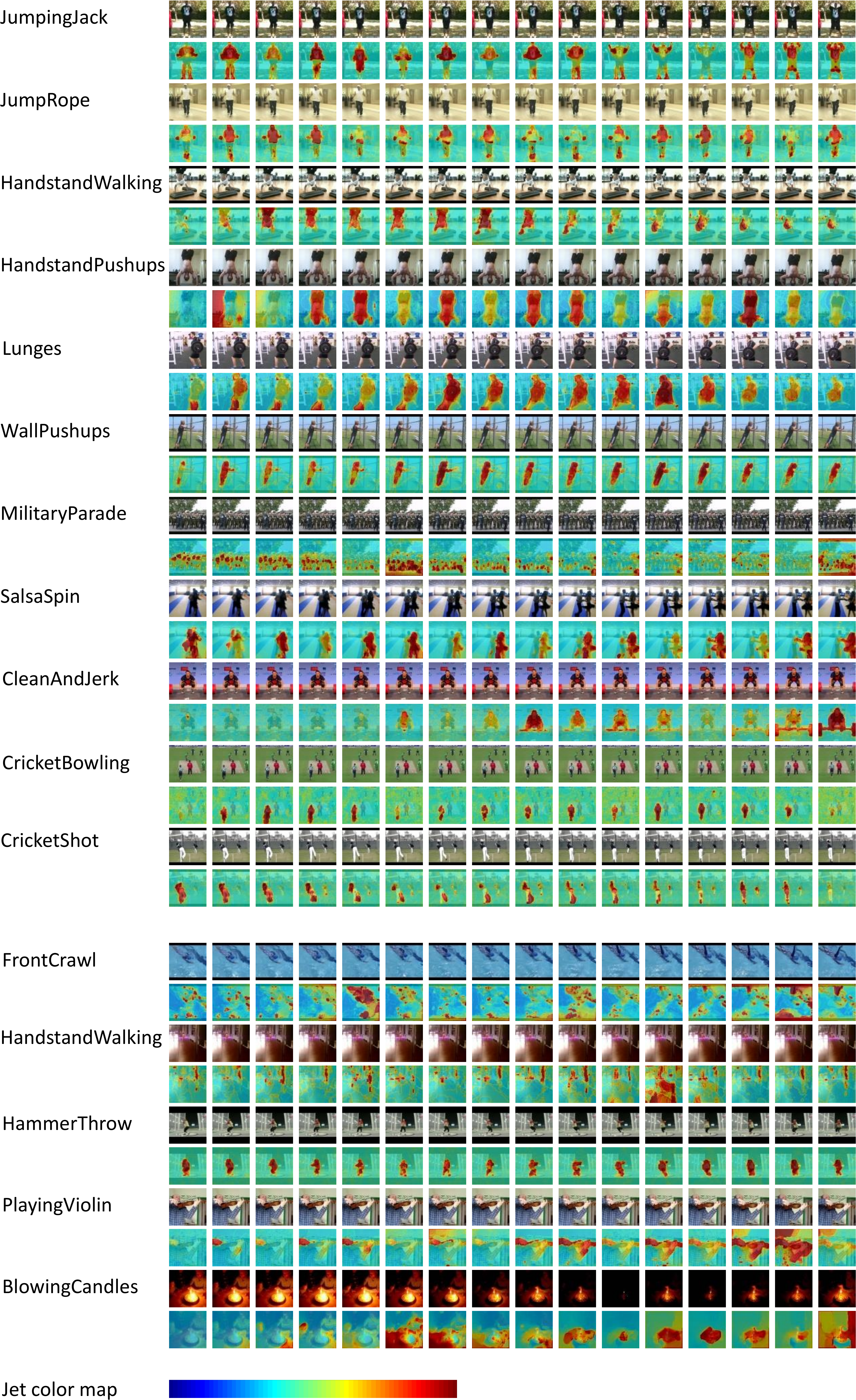}
	\end{center}
	\caption{Rows represent examples of attention over time in videos in UCF101 dataset. The top half shows examples from UCF101 of successful classes with large improvements brought about by our FCAN, while the bottom one shows examples of classes with decreases in performance. For each pair of sequences, we show original images and attention maps overlaid on images. The attention map is encoded by jet color map. The intensities are in the range [0,1], and the color scheme looks like the last row. Best viewed in color.}
	\label{fig:UCF101 attention over time}
\end{figure}

Figure~\ref{fig:UCF101 attention over time} presents some examples of flow-guided attention for selected sequences of UCF101 dataset. We show in the top half selected sequences from classes in which our FCAN model outperforms \textit{spatial-C3D}, specifically, JumpingJack, JumpRope, HandstandWalking, HandstandPushups, Lunges, WallPushups, SalsaSpin, Clean\&Jerk, CricketBowling, CricketShot. As can be observed, the FCAN model focuses on the spatio-temporally varying human torsos to make predictions. FCAN does eliminate some effects of background context by putting attention values of nearly 0.5 for background regions. We also highlight in the bottom half some cases in which our FCAN does not perform well. In these sequences, the compensated flows are erroneous due to a variety of reasons. For example, in the Front Crawl sequence, there are additional areas of focus in the swimming pool due to wave motions there that are not compensated. Similarly, in the HandstandWalking sequence, there are two distinct planes in the background which causes failure in the homography-based compensation. In the HammerThow and PlayingViolin sequences, the pertinent  motions (\eg, hand swing, bow movement) are small and/or elongated and the flow algorithm lacks the quality to clearly delineate these fine motions. Lastly, in the BlowingCandles sequence, 3D-FCAN wrongly focuses on the cake; this is due to the erroneous optical flow estimation caused by the varying candle-light illumination.

\footnotetext{The results are reproduced with our own implementations and data.}

\textbf{Ensemble of FCAN and frame-based CNN models.} Table~\ref{tab: Combination of different features} shows the results of our two-stream FCAN-comp and its ensemble with Temporal Segment Networks (TSN) \cite{Wang2016b}. The latter is essentially a two-stream frame-based CNN model; however, it takes into account frames over longer temporal range (with short snippets randomly sampled from each segment). We re-implement TSN with BatchNorm-Inception architecture for each stream of RGB and compensated flows. With our implementations, our two-stream FCAN-comp out-performs two-stream TSN \cite{Wang2016b} 0.4\% and 3.7\% on UCF101 and HMDB51 split 1 respectively. However, when we combine two kinds of features including spatio-temporal models (\ie, two-stream FCAN-comp) and frame-based models (\ie, two-stream TSN), we achieve significantly better performances. Our conjecture is that while our 3D spatio-temporal FCAN model should in principle subsume the TSN (which only randomly samples some snippets), its 3D CNN architectures may have difficulties in learning all the information, and thus there is some complementarity between the two sets of features.

\subsection{Comparison with the state of the art}
In Table~\ref{tab: Comparison with the state-of-the-art}, we compare our results to the state-of-the-art on UCF101 and HMDB51 dataset. Different methods are grouped into three categories: hand-crafted features, deep learning approaches, and attention-based networks. Note that most of methods are not directly comparable to our results because of using different network architectures and improvement schemes. Although combining hand-crafted features IDT with Fisher Vector encoding \cite{Wang2013a} is a strong baseline, our two-stream FCAN comfortably outperforms them by a margin of 6.1\% and 9.5\% on UCF101 and HMDB51 respectively. We also observe a noticeable improvement over the original two-stream 2D-CNN \cite{Simonyan2014} with 4.0\% and 7.3\% increase in UCF101 and HMDB51 respectively. We also compare to longer temporal models (\eg, LTC\cite{Varol2016}, I3D \cite{Carreira2017}), although they are not directly comparable to our work. Our temporal length is 16 frames, while they are 100 and 64 frames in LTC and I3D respectively. Our results are on par with LTC in the UCF101 dataset, but are better than LTC on HMDB51 (\ie, 66.7\% vs. 64.8\%). Two-stream I3D \cite{Carreira2017} achieves astonishing performance since they train a 3D-CNN architecture on big video dataset and fine-tune on UCF101 and HMDB51. We also achieve encouraging results compared to TSN\cite{Wang2016b} (3 modalities) on UCF101 and HMDB51, although they improve accuracy by using a better 2D-CNN architecture. In the regime of attention-based models, our method shows promising results compared to other related works. First, we outperform VideoLSTM \cite{Li2016a} by a margin of 2.8\% on UCF101 (\ie, 92.0\% vs. 89.2\%), of 10.3\% on HMDB51 (\ie, 66.7\% vs. 56.4\%). Furthermore, we also see a large margin of improvement in the performance of our two-stream FCAN model on HMDB51 when compared to that of the soft attention model \cite{Sharma2015} (\ie, 66.7\% vs. 41.3\%). We also obtain a new state-of-the-art result on Hollywood2 dataset. We attribute these successes to the explicit temporal modeling in the C3D architectures and the attentive property of the (compensated) flows.

\begin{table}
	\begin{center}
		\begin{tabular}{|l|c|c|c|}
			\hline
			 Method & UCF101 & HMDB & HW2 \\
			\hline\hline
			\cite{Wang2013a} IDT+FV 				& 85.9  &  57.2   & 64.3 \\
			\cite{Peng2014b} IDT+HSV 				& 87.9  &  61.1   & - \\
			\cite{YeLuo_ICCV_2015} IDT+Actionness	&   -   &  60.4	  & -\\
			\cite{Fernando_2015_CVPR} VideoDarwin   &   -   &  63.7   & 73.7 \\
			\cite{Fernando2016} RankPool + IDT      & 91.4  &  66.9   & 76.7 \\
			\hline
			\cite{Simonyan2014} Two-stream (avg)   & 86.9 & 	58.0 &  -\\
			\cite{Simonyan2014} Two-stream (SVM)   & 88.0 & 	59.4 &  -\\
			\cite{Ng_2015_CVPR} Two-stream LSTM    & 88.6 & 	-	 &  -\\
			\cite{Varol2016} LTC			 		& 91.7  &  64.8	  & -\\
			\cite{Wang2016b} TSN (2 modalities)		& 94.0  &  68.5   & -\\
			\cite{Wang2016b} TSN (3 modalities)		& 94.2  &  69.4   & -\\
			\cite{Carreira2017} Two-stream I3D 		& \textbf{98.0} 	&  \textbf{80.7}  & - \\
			\hline
			\cite{Feichtenhofer2016} Two-stream fusion & 92.5 &  65.4  & - \\
			\cite{Feichtenhofer2016a} ST-ResNet & 93.4 &  66.4  & -\\			
			\hline
			\cite{Sharma2015} Soft attention	& -  	&  41.3  & -\\
			\cite{Li2016a} VideoLSTM			& 89.2  &  56.4  & -\\
			\hline
			Two-stream FCAN-comp     			& 92.0  &  66.7	 &  71.1\\
			Ensemble (4 models)     			& 93.4 	&  68.2  & 	\textbf{78.4}\\
			\hline
		\end{tabular}
	\end{center}
	\caption{Comparison with the state-of-the-art on UCF101, HMDB51 and Hollywood2(HW2) with mean accuracy across 3 splits. We only compare with deep learning approaches with equal length in the temporal models, and not with handcrafted features such as IDT\cite{Wang2013a}. We would expect our results to be better after combining with IDT features.}
	\label{tab: Comparison with the state-of-the-art}
\end{table}

\section{Conclusion}
This paper introduces two-stream flow-guided convolutional attention network (two-stream FCAN) and shows that it can improve performances of the two-stream C3D. We also show that while compensated optical flows can provide some form of attention guidance, the advantage of this attention is prominent when there is explicit temporal modeling in the CNN model. The attention in our approach is modeled simply, but it shows good performances compared to the recurrent attention models for action recognition.

\clearpage

{\small
	\bibliographystyle{ieee}
	\bibliography{mybib}
}

\end{document}